\documentclass{rec2024}    % Specifies the document style.
\newdisplay{guess}{Conjecture}

\usepackage{graphicx}
\usepackage{fancyhdr}
\usepackage{algorithm}
\usepackage{algpseudocode} 
\begin{document}
\begin{article}
\begin{opening}
\thispagestyle{fancy}
\title{Interval-based validation of a nonlinear estimator}
\author{Maël \surname{Godard} ¹, Luc \surname{Jaulin} ¹, Damien \surname{Massé} ²}
\runningauthor{Maël Godard} \runningtitle{Interval-based validation of a nonlinear estimator}
\institute{¹ ENSTA Bretagne, Lab-STICC, UMR CNRS 6285, Brest, France. mael.godard@ensta-bretagne.org, luc.jaulin@ensta-bretagne.fr\\
² UBO - Université de Bretagne Occidentale, Lab-STICC, UMR CNRS 6285, Brest, France. damien.masse@univ-brest.fr}
\date{}

\begin{abstract} In engineering, models are often used to represent the behavior of a system. Estimators are then needed to approximate the values of the model's parameters based on observations.
This approximation implies a difference between the values predicted by the model and the observations that have been made. It creates an uncertainty that can lead to dangerous decision making. 
Interval analysis tools can be used to guarantee some properties of an estimator, even when the estimator itself doesn't rely on interval analysis \cite{IA_applied_to_NN_general} \cite{IA_applied_to_NN_DomOfVal}.
This paper  contributes to this dynamic by proposing an interval-based and guaranteed method to validate a nonlinear estimator. It is based on the Moore-Skelboe algorithm \cite{MS_termination}. This method returns a guaranteed maximum error that the estimator will never exceed. We will show that we can guarantee properties even when working with non-guaranteed estimators such as neural networks.

\end{abstract}

\end{opening}

\section{Introduction}
\label{Introduction}

                    % Produces section heading.  Lower-level
                    % sections are begun with similar
                    % \subsection and \subsubsection commands.

Before using any system it is important to have a model of it to predict it's behaviour. Once the modelization has been done, it is necessary to fit the model's parameter to the studied system. Estimators are used to approximate the value of these parameters based on observations made on the system. This approximation results in an error that can have a significant impact on the model if it gets too important. It can then be useful to characterise this error to validate if the estimator can be used or not.

One way to characterise it is to maximise it, i.e. to give a maximum error made by the estimator in a given context. In the case of Linear Parameter Estimation problems (LPE), many methods such as least squares have proven to be both efficient and easy to use \cite{OLS_linear_model}. These methods can be extended to solve NonLinear Parameter Estimation problems (NLPE) \cite{non_linear_LS} \cite{nonlinear_LS_estimation}, but they tend to be both less efficient and non-guaranteed, or guaranteed only in a specific context.

To overcome this problem, estimators based on Interval Analysis \cite{IA_estimator} have been developed to solve NLPE problems. They provide a set that is guaranteed to contain the true value of the parameters, rather than trying to provide their exact value. However, their use requires knowledge of Interval Analysis tools, which is not always relevant depending on the application. It is therefore important to be able to validate non-interval based estimators also in NLPE problems.

To do so, Interval Analysis tools can be applied to nonlinear and non-interval based estimators to guarantee some of their properties. For example, neural networks are often used as estimators but they fail to give a guaranteed result. Some works  have already presented applications of the Interval Analysis tools to neural networks to guarantee some of their properties, for example to evaluate their generalization \cite{IA_applied_to_NN_general}  and domain of validity \cite{IA_applied_to_NN_DomOfVal}. 

This paper contributes to this dynamic by presenting a method to provide the maximum error made by any nonlinear estimator, including neural networks. This method is able to give a guaranteed result even when the estimator itself is not guaranteed. It relies on the Moore-Skelboe algorithm \cite{MS_termination} to provide a guaranteed result. Section \ref{Notations-and-definitions} describes the notations that will later be used to formalise the problem and the method in Section \ref{Problem-formulation}. Finally, Section \ref{Application} presents an application with the validation of a gradient descent based estimator and a neural network estimator.

\section{Notations and definitions}
\label{Notations-and-definitions}

\subsection{Estimator}

An estimator is a function that approximates a quantity, or parameter, based on observed data. An estimate of a quantity \textbf{$\mathbf{q}$} is commonly denoted $\hat{\mathbf{q}}$.

Let $\mathbb{X}_{0}\subseteq\mathbb{R}^{n}$ be the set of possible
parameters and $\mathbb{E}\subseteq\mathbb{R}^{m}$ the noise set. If $\mathbf{g}:\mathbb{X}_{0}\rightarrow\mathbb{R}^{m}$ is the observation function and $\mathbf{\mathbf{\psi}}:\mathbb{R}^{m}\rightarrow\mathbb{X}_{0}$ is
the estimator to validate, the estimation problem can be defined for all $\mathbf{x} \in \mathbb{X}_{0}$ and $\mathbf{e} \in \mathbb{E}$ by :
\begin{eqnarray}
  \mathbf{y} & = & \mathbf{g}(\mathbf{x})+\mathbf{e}   \label{noisy_obs}\\
  \hat{\mathbf{x}} & = & \mathbf{\mathbf{\psi}}\left(\mathbf{y}\right) \label{estimation}
\end{eqnarray}
Where (\ref{noisy_obs}) is the Noisy observation and (\ref{estimation}) is the estimation.

\begin{figure}[H] % figure 1
\centering
\includegraphics[scale=0.7]{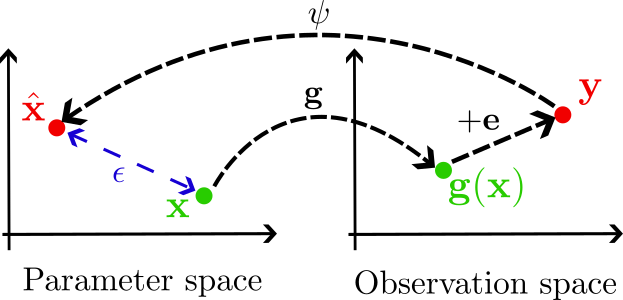}\\
\caption[]{Estimation problem} \label{Estimation-problem}
\end{figure}

We also define the error function $\epsilon:\mathbb{X}_{0} \times \mathbb{E}\rightarrow\mathbb{R} $  which gives the error of the estimator for a given parameter vector $\mathbf{x}$ and a given noise $\mathbf{e}$.
\begin{equation}
    \epsilon(\mathbf{x},\mathbf{e})=\|\mathbf{x}-\hat{\mathbf{x}}\|=\|\mathbf{x}-\psi\left(\mathbf{g}\left(\mathbf{x}\right)+\mathbf{e}\right)\|
\end{equation}
Figure \ref{Estimation-problem} represents these notions in two different spaces: the parameter space and the observation space. Starting from the real value $\mathbf{x}$, an observation $\mathbf{y}$ is made. It can be decomposed into a perfect observation $\mathbf{g(x)}$ and a noise $\mathbf{e}$. The estimator finally relies on this observation to provide the estimated value $\hat{\mathbf{x}}$. The error $\epsilon$ represents the distance between $\mathbf{x}$ and its estimate $\hat{\mathbf{x}}$.

To validate an estimator we want to evaluate the maximum of its error function on $\mathbb{X}_{0}\times\mathbb{E}$, noted $\bar{\epsilon}$. The problem can written as below:
\begin{eqnarray}
  \bar{\epsilon} & = & \max \epsilon (\mathbf{x},\mathbf{e}) \label{maxim_pb} \\
  \mathbf{x} & \in & \mathbb{X}_{0} \\
  \mathbf{e} & \in & \mathbb{E}
\end{eqnarray}

\subsection{Intervals and boxes}

Intervals are used to represent a range of value instead of a single one. An interval can be defined by two of the following:

\begin{itemize}
\item Its lower bound, or left bound, noted $lb$
\item Its upper bound, or right bound, noted $ub$ or $rb$
\item Its width, noted $w$
\end{itemize}

For a given interval $I$, the parameters described above are linked by the relation :
\begin{equation}
    rb\left(I\right)=lb\left(I\right)+w\left(I\right)
\end{equation}
The cartesian product of two or more intervals is called a box, and the set of n-dimensionnal boxes is denoted $\mathbb{IR}^{n}$ for $n\geqslant1$. For any function $\mathbf{f}:\mathbb{R}^{n}\rightarrow\mathbb{R}^{m}$, its definition can be extended to $\mathbb{IR}^{n}$ and we also call $\mathbf{f}:\mathbb{IR}^{n}\rightarrow\mathbb{IR}^{m}$ its natural inclusion function.

These boxes can for example be used to form a cover of a subset of $\mathbb{R}^{n}$. A cover of a set $\mathbb{X}$ is a family of subsets of $\mathbb{X}$ whose union is all of $\mathbb{X}$. If $C=\left\{ U_{\alpha}:\alpha\in A\right\} $ is an indexed family of subsets $U_{\alpha}\subset\mathbb{X}$, then C is a cover of $\mathbb{X}$ if $\bigcup_{\alpha\in A}U_{\alpha}=\mathbb{X}$ \cite{cover_book}.

\subsection{Optimization}

An Optimization problem, also called mimization problem, aims to find the minimum of a given function. It is defined by :

\begin{itemize}
    \item An objective function to minimize $f:\mathbb{R}^{n}\rightarrow\mathbb{R}$
    \item A domain $\mathbb{X}_{0}\subseteq\mathbb{R}^{n}$
    \item In the case of a constraint optimization problem, a set of conditions
$g_{i}\left\{ x_{1},\ldots,x_{n}\right\} \leq0$ for $i\in\left\{ 1,\ldots,m\right\} $
where $g_{i}$ are functions of type $\mathbb{R}^{n}\rightarrow\mathbb{R}$.
\end{itemize}

Algorithms to solve this kind of problem are called Optimization algorithms. Among them, Global Optimization algorithms focus on finding the global minimum of the objective function, noted $\mu$ below.

\subsection{Moore-Skelboe Algorithm}

A classical Global Optimization algorithm is the Moore-Skelboe algorithm. This algorithm takes as inputs the objective function $f$, its domain $\mathbb{X}_0$ and a stopping criterion $\delta$. It returns an interval containing the global minimum of the objective function, which bounds are an optimistic and a pessimistic approximation of the global minimum. Multiple variations of the algorithm exist \cite{MS_termination}, Algorithm \ref{MS0} displays its basic implementation.

\begin{algorithm} [htb]
\begin{algorithmic}
    \State Let $\left\{ B_{0}\right\} $ be a cover of $\mathbb{X}_0$
    \While{$w\left(f\left(B_{0}\right)\right)>\delta$}\Comment{stopping criterion to choose}
        \State Remove $B_{0}$ from the cover
        \State Split $B_{0}$ 
        \State Insert the result into the cover in increasing order 
        \State of $lb\left(f\left(B_{i}\right)\right)$, for $i=\{0,\ldots,N-1$\}
    \EndWhile
    \State \textbf{return} $f\left(B_{0}\right)$ \Comment{$\mu \in f(B_0)$}
\end{algorithmic}
\caption[]{Moore-Skelboe algorithm (${MS}_{0}$)} \label{MS0}
\end{algorithm}
    
When implementing this algorithm it is also possible to return the last $B_0$ to obtain a box containing an antecedent of a value between $\mu$ and $\mu + \delta$.

\begin{figure}[htb] % figure 1
\centering
\includegraphics[scale=0.4]{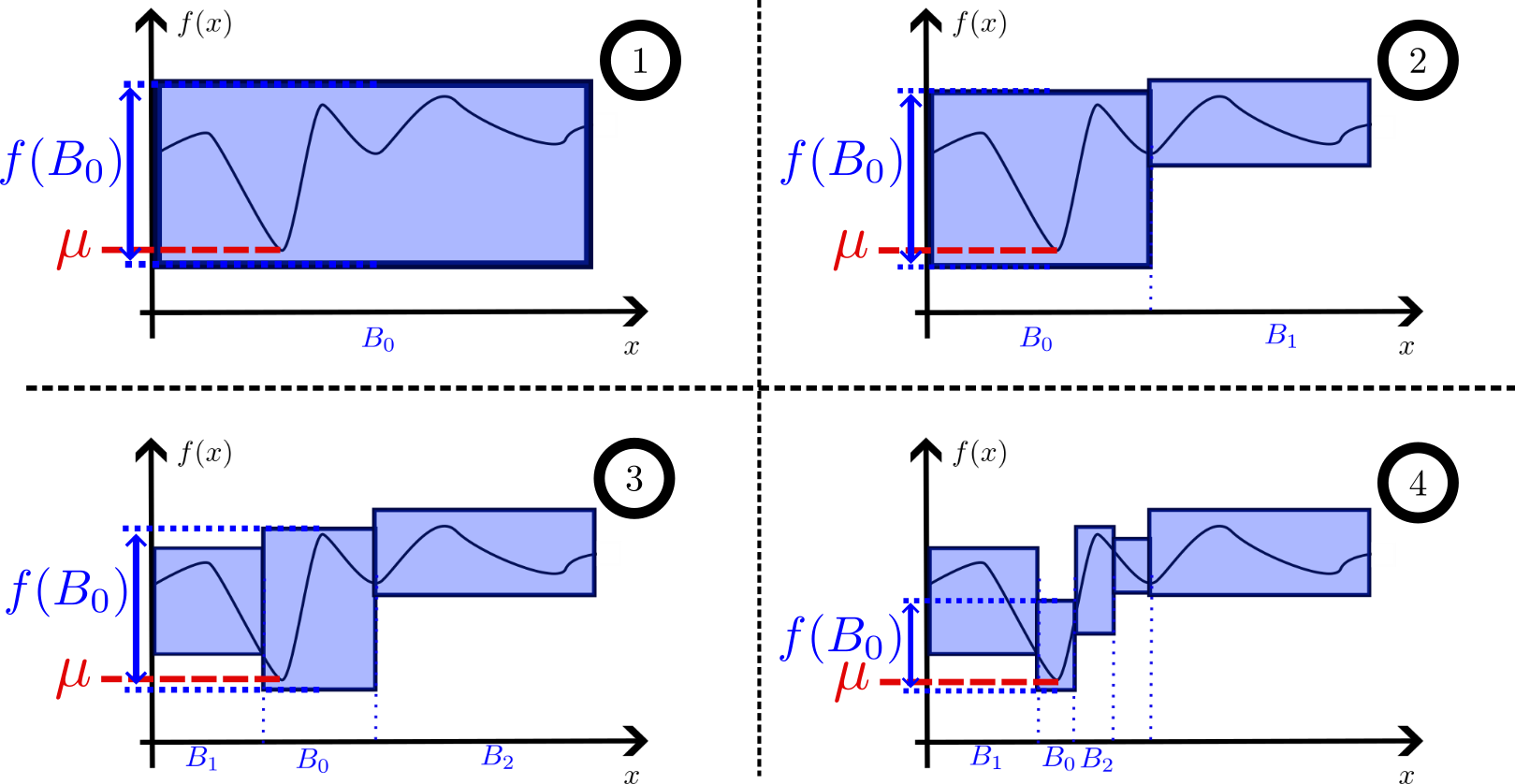}\\
\caption[]{Moore-Skelboe algorithm example} \label{MS_example}
\end{figure}

The logic behind this algorithm is that at each iteration the box $B_0$ is chosen so that the interval $f(B_0)$ is the $f(B_i)_{i\geq 0}$ interval with the lower left bound. As $\{B_0,B_1,...,B_{N-1}\}$ covers the domain $\mathbb{X}_{0}$:
\begin{equation}
    \label{mu_geq}
    \forall x \in \mathbb{X}_{0}, lb(f(B_0)) \leq f(x)
\end{equation}
From this relation we deduce:
\begin{equation}
    \label{mu_geq}
    lb(f(B_0)) \leq \mu
\end{equation}
As $B_0$ is not an empty box, $f(B_0)$ always contains at least one value. By definition of the global minimum of a function:
\begin{equation}
    \label{mu_leq}
    \mu \leq rb(f(B_0))
\end{equation}
Finally from \ref{mu_geq} and \ref{mu_leq}:
\begin{equation}
    \label{mu_in}
    \mu\in f(B_0)
\end{equation} 
This result is illustrated Figure \ref{MS_example}.

\section{Problem Formulation and method}
\label{Problem-formulation}

To validate an estimator, we want to estimate the maximum error that it commits on a given set of possible parameters. This maximisation problem on the function $\epsilon$ can also be interpreted as a Optimization problem on the function $-\epsilon$.

The Moore-Skelboe algorithm can then be applied with the objective function $f=-\epsilon$ and with $\{B_0=\mathbb{X}_0\times \mathbb{E}\}$ as the initial cover. 

Upon splitting $B_0$, only its component belonging to $\mathbb{X}_0$ will be splitted. By doing so the algorithm can also give the information of which parameter value is the most sensitive to noise when performing the estimation.

Given the context, we want to ensure that our estimation of the maximum error is pessimistic if not exact, so the focus will be on $-lb(f(B_0))$.

\section{Application}
\label{Application}

\subsection{Application case}

\begin{figure}[htb] % figure 1
\centering
\includegraphics[scale=0.65]{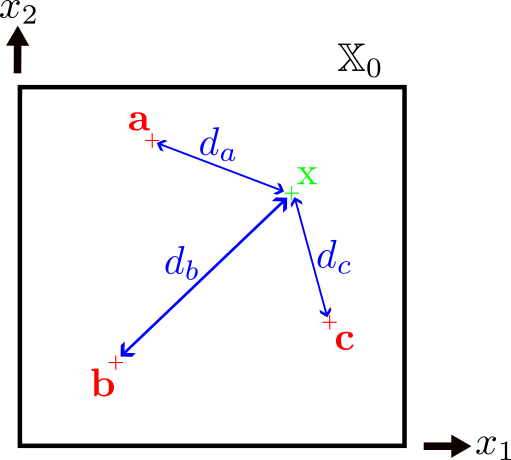}\\
\caption[]{Application example} \label{appli_ex}
\end{figure}

To illustrate an application of this method the case
described Figure \ref{appli_ex} will be considered. In this case the
parameter to estimate is the 2-Dimensional position noted $\mathbf{x}$,
and the observation is the distances between x and three landmarks
noted \textbf{$\mathbf{a}$}, \textbf{$\mathbf{b}$} and $\mathbf{c}$.
The observation function is then $\mathbf{g}:\mathbb{R}^{2}\rightarrow\mathbb{R}^{3}$, defined
by :
\begin{equation}
\mathbf{g}\left(\mathbf{x}\right)=\left(\begin{array}{c}
d_{a}\\
d_{b}\\
d_{c}
\end{array}\right)=\left(\begin{array}{c}
\|\mathbf{a}-\mathbf{x}\|\\
\|\mathbf{b}-\mathbf{x}\|\\
\|\mathbf{c}-\mathbf{x}\|
\end{array}\right)
\end{equation}

The norm used below to calculate the distances is the Euclidian norm. For numerical applications, we will take :
\begin{itemize}
    \item $\mathbb{X}_{0}=\left[5,25\right]^{2}$
    \item $\mathbb{E}=\left[-0.2,0.2\right]^{3}$
    \item $\ensuremath{a=\left(10,-9\right)}\ensuremath{;b=\left(5,12\right)};\ensuremath{c=\left(-15,0\right)}$
\end{itemize}

\subsection{Neural network estimator}

\begin{figure}[htb] % figure 1
\centering
\includegraphics[scale=0.55]{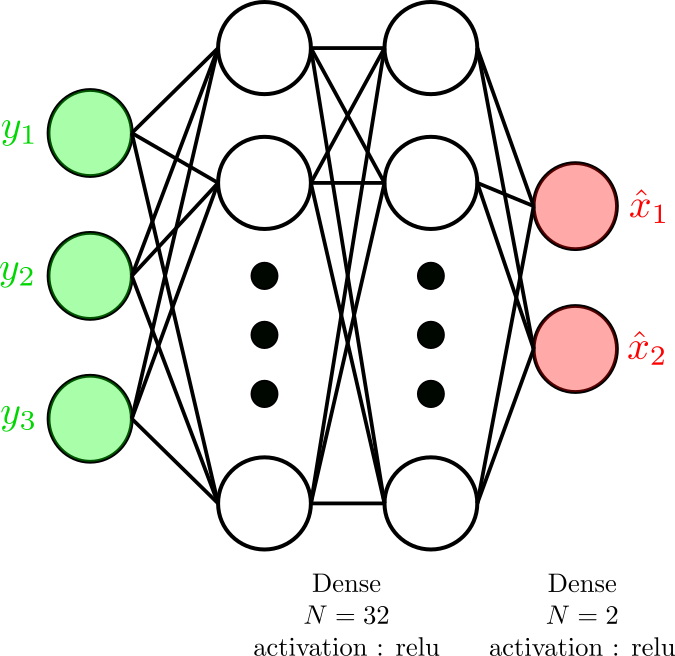}\\
\caption[]{Neural network estimator} \label{nn_estimator}
\end{figure}

Neural networks are a common example of non-guaranteed estimator.
To prove the efficiency of our method, we created a neural network
estimator represented Figure \ref{nn_estimator}.
It is composed of :
\begin{description}
\item[An Input layer] of 3 neurones for the three noisy distances.
\item[Two Denses Hidden layers] of 32 neurones with relu activation.
\item[A Dense Ouput layer] of 2 neurones with relu activation for the two components of $\hat{\mathbf{x}}$
\end{description}

The relu function can be defined on $\mathbb{R}$ by $relu(x)=max(0,x),x \in \mathbb{R}$. The relu activation means that in each layer the relu function will be applied on the output, making the neural network estimator nonlinear.
%ajoute que IA ajoute pessimisme et rappeler que NN non garanti

This neural network has then been trained on noisy data. The estimation is finally obtained by making a prediction with the resulting trained neural network.

\subsection{Numerical application}

\begin{figure}[htb] % figure 1
\centering
\includegraphics[scale=0.7]{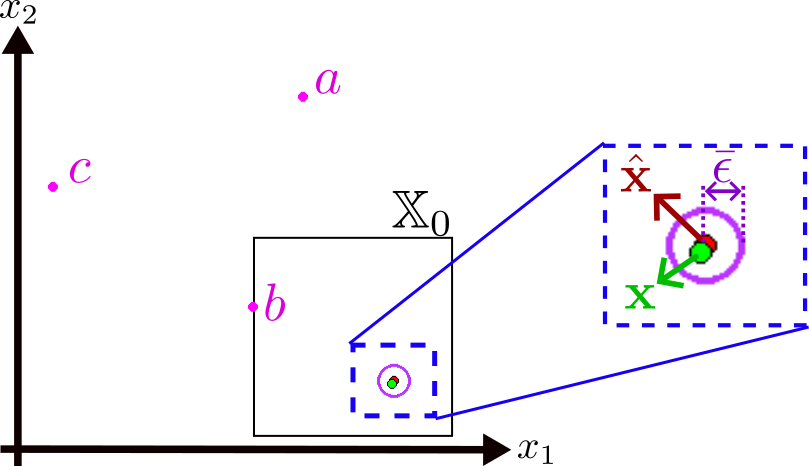}\\
\caption[]{Simulation to visualise $\bar{\epsilon}$} \label{simul_eps}
\end{figure}

The maximum error given by the Moore-Skelboe algorithm in this case is \textbf{$\bar{\epsilon}=1.7$}. A small simulation, shown in Figure \ref{simul_eps}, allows us to visualise this $\bar{\epsilon}$.

 For this simulation, noisy measurements are generated and then passed to the estimator. The result, in red, can then be compared to the expected value, in green. As $\bar{\epsilon}$ represents the maximum distance between  $\mathbf{x}$ and  $\hat{\mathbf{x}}$, the green dot representing $\mathbf{x}$ should always be in the purple circle of radius $\bar{\epsilon}$ centered at $\hat{\mathbf{x}}$. 

 Depending on the accuracy required by the application, we can then either use the neural network as it is or train it again, possibly by modifying its structure.
 
 Using this method, we were able to find a guaranteed property for a neural network estimator that is both non-guaranteed and nonlinear. As the neural network presented here was simple the basic implementation of the Moore-Skelboe algorithm was sufficient. For more complex estimator variants of the algorithm may prove more efficient as the one presented here gives an increasingly pessimistic result as the estimator becomes complex.

%TODO ajouter le fait que plus c'est complexe plus on est pessimiste

\section{Conclusion}

This paper proposes an interval-based and guaranteed method to validate a nonlinear estimator thanks to the Moore-Skelboe algorithm. It allows us to find a guaranteed property
in the form of a maximum error even when working with non-guaranteed estimators such as neural networks. Variants of the Moore-Skelboe algorithm presented in \cite{MS_termination} could be
implemented to make this method more efficient in terms
of time and computational complexity, but the application
presented here shows that the basic implementation of the
Moore-Skelboe algorithm is sufficient in simple cases.

\acknowledgements

This work has been supported by the Brittany region (ARED).

\end{article}
\end{document}